\documentclass[journal]{IEEEtran}
\usepackage{microtype}
\usepackage{graphicx}
\usepackage{subfigure}
\usepackage{booktabs} 
\usepackage{hyperref}
\usepackage{caption}
\usepackage{listings}
\usepackage{array}
\usepackage{changepage}

\usepackage{algorithmicx}
\usepackage{algorithm}
\usepackage{algpseudocode}
\usepackage{algpseudocode}
\usepackage{amsmath}
\usepackage{placeins} 
\usepackage{multicol}
\ifCLASSINFOpdf
\else
\fi

\begin{document}

\title{Agentic Retrieval and Reinforcement Learned Equation Chains: A Controlled Generation Framework for Complex and Novel Physics Word Problems}

\author{Tirthankar~Mittra
        % \thanks{Manuscript received April 19, 2005; revised August 26, 2015.}
        }

\markboth{Journal of \LaTeX\ Class Files,~Vol.~14, No.~8, August~2015}%
{Shell \MakeLowercase{\textit{et al.}}: Bare Demo of IEEEtran.cls for IEEE Journals}

\maketitle

\begin{abstract}
Generating high quality Physics Word Problems(PWPs) that exhibit complexity, novelty, and improved solvability represents a significant, yet under researched, challenge in educational and AI-driven content generation. Existing methodologies, including those adapted from the more explored domain of Math Word Problems(MWPs), frequently fail to produce robust content, these approaches often result in problems that are mathematically unsolvable or ambiguous, overly simplistic in structure, or constrained by limited linguistic and conceptual complexities. In this paper, we introduce a novel, two stage generative framework called ARVRE (Agentic Retrieval Value Reinforced Equation-chain), designed to overcome these fundamental limitations. The core of our innovation is driven by offline temporal difference updates of the action value function, that programmatically links multiple valid physics equations, and an agentic RAG framework that dynamically selects topic words for a physics word problem. Our approach also facilitates fine grained control over problem difficulty. In the second stage, we leverage the power of Large Language Models (LLMs) to translate the equations and topic phrases into a physics question. By using the equation chain and topic words as an explicit, constrained prompt, we guide the LLM to maximize novelty while retaining the guaranteed mathematical correctness. Through rigorous human and automated evaluations, we demonstrate that our framework achieves significant improvements across several key metrics. Our generated PWPs show a marked increase in complexity (problems requiring more conceptual steps to solve), novelty (moving beyond standard textbook templates), and correctness (better solvability guarantees). This work establishes our method as a reliable and promising tool for automatically creating novel and challenging physics content suitable for both educational resource development and advanced research in generative AI.
\end{abstract}

% Note that keywords are not normally used for peer-reviewed.

% For peer review papers, you can put extra information on the cover
% page as needed:
% \ifCLASSOPTIONpeerreview
% \begin{center} \bfseries EDICS Category: 3-BBND \end{center}
% \fi
%
% For peer-review papers, this IEEEtran command inserts a page break and
% creates the second title. It will be ignored for other modes.
\IEEEpeerreviewmaketitle

\section{Introduction}

\IEEEPARstart{P}{hysics} word problems (PWPs) play a vital role in education, helping students improve their proficiency and understanding of fundamental concepts. For teachers, PWPs serve as invaluable tools for assessing student competency, allowing teachers to gauge how well students grasp complex ideas. These problems help educators determine whether a student has truly understood the concepts taught or whether adjustments to their teaching strategies are necessary to better meet the needs of the class. Each PWP is constructed around a core set of equations that students must identify and solve, requiring them to engage in both analytical thinking and effective problem-solving skills. 
In recent years, teachers have faced increasing challenges in generating original physics questions due to the widespread availability of solutions online, making it easy for students to plagiarize during assignments or exams \cite{mccabe2012cheating}. Furthermore, tools such as ChatGPT can easily solve standard physics problems, further complicating the issue \cite{ventayen2023chatgpt}. In response, educators often spend considerable time creating new questions and ensuring that they are not easily searchable. Although this effort is essential, it can place a significant burden on teachers, detracting from the time they could devote to improving their pedagogical methods and fostering student engagement.
The Physics Word Problem (PWP) generator addresses this challenge by automatically generating unique and complex problems. This not only saves time for teachers but also provides students with challenging questions to improve their problem-solving skills, conceptual understanding, and mental acuity. \cite{buteler2016solving} found that engaging students in problem-solving helps them develop a deeper understanding of physical principles by connecting mathematical formalism with conceptual reasoning. 

Deep learning has made significant progress in natural language processing(NLP) tasks like Text Generation\cite{see2017get}\cite{yu2022survey}, Sentiment Analysis\cite{kenton2019bert} \cite{mukherjee2013sentiment}, Text Classification\cite{chen2015convolutional}, Text Summarization\cite{see2017get}, Machine Translation\cite{vaswani2017attention} \cite{garg2018machine}, Question Answering\cite{rajpurkar2016squad} \cite{soares2020literature} etc, likewise we wanted to study it's efficacy in generating word problems, especially with the advent of modern causal language models. Traditionally, the generated problems had issues, like being unsolvable or being topically irrelevant. Figure [\ref{fig:probEq}] illustrates a few such issues.
\\
\\
\textbf{Key Contributions}:  The \textit{ARVRE} algorithm we designed improves the quality of the generated physics word problems by improving four important characteristics of it, and those are  \textit{complexity},  \textit{novelty},  \textit{solvability}, and  \textit{generating user-aligned problems}. Our algorithm has two main components, the first component is the \textbf{Equation Builder}, which intelligently constructs a set of equations with the corresponding set of known and unknown variables that form the mathematical foundation for our physics word problem. The second component is the \textbf{Topic Phrase Selector}, which retrieves relevant topic phrases for the selected equations using an agentic RAG(retrieval augmented generation)\cite{singh2025agentic}  approach. Both the equations and topic phrases are subsequently fed to the LLM, which generates the final word problem. A value-based reinforcement learning is then used to reinforce high-quality equation sets for the given LLM. We have specifically used a modified SARSA \cite{rummery1994line} algorithm with offline forward TD(0) updates to propagate the reward, after the end of each episode. Each of these components is described in greater detail in the next section.
\\
\\
We now discuss how our algorithm improves the four characteristics of a physics problem.

\textbf{(1) Complexity:} Standard LLMs use only a few equations per problem. Our method builds word-problems by chaining multiple equations, making them richer and more challenging to solve.

\textbf{(2) Novelty:} Our system \textbf{ARVRE} is capable of selecting all valid equation combinations, covering the full problem space. It then injects topic-specific language using a novel RAG\cite{singh2025agentic} solution. Finally the equation sets that produce novel problems are reinforced through SARSA based learning and so our framework keeps getting better over time.

\textbf{(3) Solvability:} LLMs can't guarantee a problem is actually solvable. We improve this by ensuring every equation system has N equations for N unknowns, validating them after generation, and rewarding sets that work, this makes reliable problems more likely to reappear.

\textbf{(4) Generating user aligned problems:} Users can personalize word problem difficulty directly via a threshold and reward function, this allows users to match their own learning needs.
\\
\\
The source code and additional resources can be found at \url{https://github.com/tirthankar95/agentic-physics-question-generator.git}.

\begin{figure}
\centering
    \begin{tabular}{|m{7.5cm}|c|}
        \hline
        \\
         Physics Equations (Simple Kinematics)\\
         \\
         $v = u + a \times t$\\
         $F = m \times a$\\
         $S = u \times t + \frac{1}{2}a \times t^{2}$\\
         \\
         \hline
         \\
         Generated Physics Questions\\
         \\
         \textbf{Unsolvable Question}: A person throws a 1 kg ball with an initial velocity of 10 $m/s$. Find the acceleration and force applied to the ball initially and the distance covered by the ball after 10 seconds.\\
         \\
         \textbf{Simple Question}: A force of 10 N is applied on a car of 1000 kg, find the acceleration of the car.\\
         \\
         \hline
    \end{tabular}
     \caption{Examples of bad problems generated by typical Physics Word Problem generators given the associated equations from the selected topic.}
     \label{fig:probEq}
\end{figure}

\section{Related Work}
Research on generating Physics Word Problems(PWPs) has been largely overlooked. Since the 1970s, significant efforts have been made to develop systems that solve physics problems rather than generate them. Early approaches, such as \cite{novak1977representations}, focused on breaking down English descriptions of physics problems into different structured representations, including canonical forms and geometric models, to facilitate problem-solving. More recent works, such as \cite{leszczynski2016machine} and \cite{bleiweiss2019neural}, have continued this trend by solving PWPs with limited physical models, such as free-falling objects. In contrast, \cite{ding2023using} demonstrated that Large Language Models (LLMs) can solve a much broader range of physics problems while providing proper explanations. The authors of the paper curated a dataset to demonstrate the power of LLMs. While these studies showcase advancements in solving physics problems, little attention has been given to the generation of diverse and complex PWPs.

In comparison, the Math Word Problem(MWP) generation(a very similar problem) has received significantly more attention. However, existing methods face several key limitations, as illustrated by the following examples.
Template-based approaches, such as \cite{polozov2015personalized}, rely on manually crafted, rigid structures that constrain both the linguistic diversity and mathematical complexity of generated problems. Extending these systems requires extensive manual effort, including writing problem logic in declarative languages like Answer Set Programming (ASP)\cite{lifschitz2002answer}, which is non-trivial for users unfamiliar with formal logic.
Methods that use complex architectures and math expression trees, like in \cite{wu2022automatic} \cite{polozov2015personalized}, are typically limited to generating problems based on a single equation(effectively), which inherently restricts the complexity of the output. Moreover, in systems like \cite{wu2022automatic}, generation is influenced by a problem solver that initially can only solve simple problems. This creates a feedback loop that encourages the model to generate similarly simple questions.
Additionally, approaches based on RNN models such as GRUs, BiLSTM \cite{zhou2019towards} \cite{wu2022automatic}, as opposed to LLMs, tend to generate problems that are not only structurally simplistic but also linguistically constrained. On the other hand, methods like \cite{wang2021math} fine-tune GPT-2 using a combination of equation consistency loss and token-level cross-entropy loss, which makes the process complicated due to the challenges in curating high-quality datasets. Additionally, the consistency generator itself is based on GPT-2, which is not explicitly trained to extract equations from math word problems (MWPs). Since only the problem generator module is trained, inconsistencies may arise in enforcing equation consistency loss. Our approach differs fundamentally; instead of curating large datasets and finetuning LLMs—an expensive and computationally intensive process—we first generate a valid set of equations and then use LLMs solely to generate the surrounding text. This method is not only simpler but also more robust at generating novel, complex questions.

While we draw inspiration from existing Math Word Problem MWP) generation methods, these approaches often fall short of producing problems with sufficient mathematical complexity and linguistic diversity. Our method addresses these limitations by first constructing more complex equations and then using large language models (LLMs) to generate diverse, conceptually rich Physics Word Problems.

\section{Methodology}
\begin{figure}
    \captionsetup{type=lstlisting} 
    \centering

%%% ALGORITHM 1 %%%
\begin{minipage}{0.5\textwidth}    
\begin{algorithm}[H]
\caption{Physics Word Problem Generator (main routine)}
\begin{algorithmic}[1]
\Procedure{PhysicsProblemGenerator}{\texttt{topic}}
\State Load:
\State \quad - equations for the given topic
\State \quad - LLM configuration
\State \quad - action-value function $Q(s,a) \forall s,a$

\State Construct a graph $G$ using the topic equations

\Repeat
    \State Select $S_{eqn} \subseteq$ known/unknown variables from $G$
    \State $S_{eqn} \gets$ BuildEquationSet(topic, $G$)
    \State Create template based prompt $P_{eqn}$ from $S_{eqn}$
\Until{$S_{eqn}$ is valid}
\State Extract topic words $\tau$ from the equation prompt $P_{eqn}$
\State $\tau \gets ExtractTopicWords(P_{eqn})$
\State Replace words in $\tilde{\tau}$ with synonyms to get $\tau$
\State Generate a question $q \gets$ using LLM $\mathcal{L}(P_{eqn}, \tilde{\tau})$
\State Refine $q$ using $\mathcal{L}$ to ensure validity, producing $\tilde{q}$
\State Get Reward $r \gets$ from LLM + Human Judge
\If{SARSA training is enabled}
\For{$(s, a, r, s', a') \in$ $G$ traversal}
    \State $Q(s, a) \leftarrow (1-\alpha)Q(s, a) + \alpha \, (r + \gamma \, Q(s', a'))$
\EndFor
\EndIf
\State \Return $\tilde{q}$
\EndProcedure
\end{algorithmic}
\end{algorithm}
\end{minipage}
%%% ALGORITHM 1 %%%

%%% ALGORITHM 2 %%%
\begin{minipage}{0.5\textwidth}
\begin{algorithm}[H]
\caption{How to select known \& variables from Graph $G$}
\begin{algorithmic}[1]
\Function{BuildEquationSet}{topic, Graph $G$}

\State Sample a starting equation $e_0$ from the topic
\State Initialize visited set $V_{vis} \gets \emptyset$

\Function{GetUnknowns}{$e$}
    \State Initialize unknown variables set $U \gets \emptyset$
    \State Mark $e$ as visited: $V_{vis} \gets V_{vis} \cup \{e\}$
    \State E $\gets$ Get all edges from equation $e$
    \If{$\text{rand}() \geq TH$}
        \State Sample edge $a \sim \text{Uniform}(E)$
        \If{$\text{rand}() \leq 1 - \epsilon$}
            \State Exploit Q values: $a \gets \arg\max_{a'} Q(e, a')$ 
        \EndIf

        \State $e' \gets$ Next equation node from $G$ given $(e, a)$ 
        \If{new edge $e'$ not in $V_{vis}$}
            \State $U_e' \gets \Call{GetUnknowns}{e'}$
            \State Accumulate unkowns: $U \gets U \cup a \cup U_e'$
        \EndIf
    \EndIf

    \State \Return $U$
\EndFunction

\State $U \gets \Call{GetUnknowns}{e_0}$ 
\State $e_{last} \gets$ Get last equation from graph $\Gamma$ traversal 
\State $E_l \gets$ Get a set of all variables in $e_{last}$ but $\notin U$
\State Add additional unknown: $U \gets U \cup \text{Uniform}(E_{l})$

\State Compute known variables: $K \gets V - U$

\State \Return $K$ (known), $U$ (unknown)

\EndFunction
\end{algorithmic}
\end{algorithm}
\end{minipage}
%%% ALGORITHM 2 %%%

\caption{Physics Word Problem Generator Algorithm}
\label{alg:completeProcess}
\end{figure}

The task of generating Physics Word Problems can be represented by Equation~\ref{PWP_prob}, which is the product of the probability of generating each token $s_t$, conditioned on the set of valid equations along with their known and unknown set($\mathbf{X}^e$), plus the set of relevant topic phrases ($\mathbf{X}^{tp}$), and all previously generated tokens $s_{<t}$. 

\begin{equation}
p_{\theta}(P \mid X^e, X^{tp}) =
\prod_{t=1}^{T} p_{\theta}(s_t \mid X^e, X^{tp}, s_{<t})
\label{PWP_prob}
\end{equation}

\begin{figure*}
\centering
\includegraphics[width=1\linewidth]{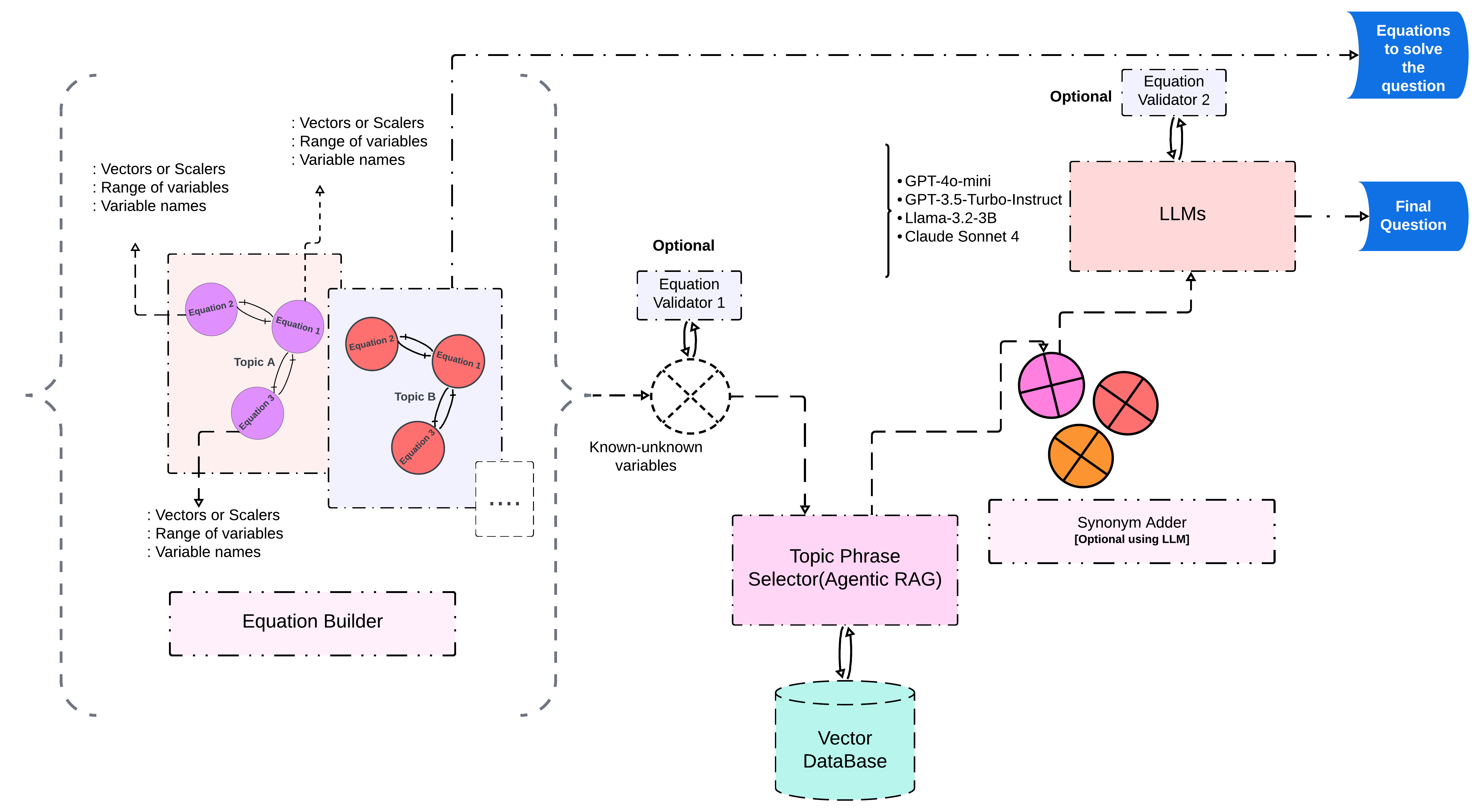}
\caption{Overview of our Physics Word Problem Generator (PWPGen) network.}
\label{fig:arch}
\end{figure*}

% \begin{figure}
% \includegraphics[width=0.5\textwidth]{Algo_main.png}
% \caption{Examples of topics, intermediate prompts, and LLM-generated final response from left to right column, using our approach. Topic phrases in intermediate prompts are redacted.}
% \end{figure}

\begin{figure}
\centering
\includegraphics[width=0.7\linewidth]{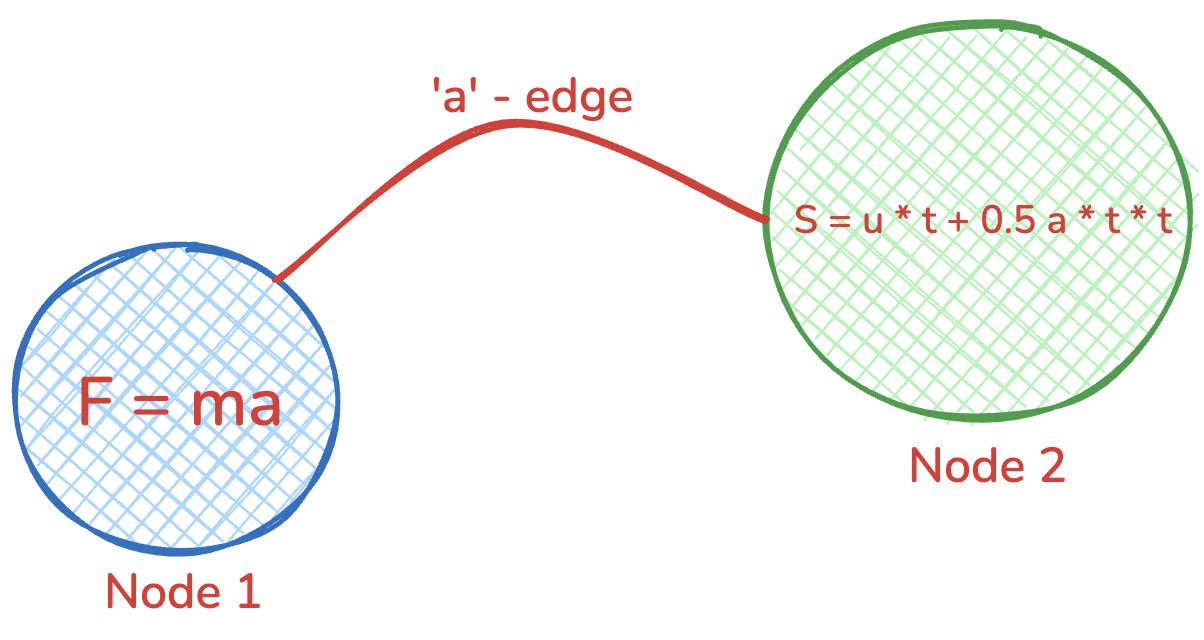}
\caption{A toy equation graph constructed from the Simple Laws of Motion topic, illustrating the relationships between equations(nodes) through a shared variable(edge).}
\label{fig:toy}
\end{figure}

Our Physics Word Problem Generator consists of five key components, illustrated in Figure~\ref{fig:arch}: \textbf{(1) Equation Builder}, \textbf{(2) Topic Phrase Selector}, \textbf{(3) Synonym Adder}, \textbf{(4) LLMs}, and \textbf{(5) Equation Validator (optional)}. The complete algorithm outlined in the Listing~\ref{alg:completeProcess}, \textbf{Algorithm 1}, is the main entry point of the code. \\
\\
To explain our framework, we use a toy example in the following section. Physics questions are generated topic-wise, where each topic is associated with a curated set of equations. Every generated physics question is based on an underlying set of equations selected from this collection. When a user requests a question from a particular topic, the \textbf{Equation Builder} component selects a subset of equations from the set of all equations associated with that topic. Let’s take a step backwards, even before the user asks our framework to generate a question, the \textbf{Equation Builder} component pre-computes a graph $G(u, v)$ for each topic, where $u$ denotes the set of equation nodes and $v$ denotes the set of undirected edges. Each node represents an equation, and an edge connects two equations that share a common variable. For example, let’s say the user selects “Simple Laws of Motion” as the topic, and let’s say the topic has the following set of equations.
$$S = ut + \frac{1}{2}at^2 \qquad \text{toy-equation (1)}$$
$$F = ma \qquad \text{toy-equation (2)}$$
For our example, the graph will have two equation nodes, one for toy-equation(1) and the other for toy-equation(2). These equations are connected by the undirected edge $(a)$ as shown in Figure~\ref{fig:toy}. If two equations share multiple variables, they will be connected by multiple edges, each labeled with the corresponding shared variable.

The process of generating an equation set begins by sampling a random seed equation node from a given topic. From this initial state $s$ (equation node), we expand to a neighboring (child) node by selecting an outgoing edge $a$ that increases the likelihood of maximizing the action-value function $Q(s,a)$. Think of the action-value function $Q(s, a)$ as the expected reward when action $a$ is selected from the current state $s$, where $s$ represents the current equation node and action $a$ is the edge, i.e., the common variable that connects two equation nodes. Essentially, we are using reinforcement learning to guide which equations to select for a particular topic, during training the reward comes from a combination of using LLM as a judge and user feedback on the final generated physics question based on parameters like solvability and novelty. Details surrounding the training of our reinforcement learning algorithm is present in the next subsection (A. Equation Builder). The expansion from the starting state $s$ (equation node) follows an $\epsilon$-greedy policy \cite{sutton1998reinforcement}, where with probability $1 - \epsilon$, the greedy action $argmax_a Q(s,a)$ is selected, and with probability $\epsilon$, an action is chosen uniformly at random from all the $N$ outgoing edges. Therefore, the total probability of selecting the greedy action is $1-\epsilon+\epsilon/N$, and for the rest of the $N-1$ actions, the individual probability of selection will be $\epsilon/N$. Continuing our example, let's say the starting equation that was randomly selected was $F = ma$, then we can travel to the next equation node via $(a)$ given the probability of transition is greater than the threshold $P(\theta) > TH$ and the $\epsilon$-greedy policy returned $(a)$, the probability of transition $P(\theta)$ is used to control early graph termination and $TH$ is known as the difficulty threshold, lowering the threshold would mean that a more complicated set of questions are generated. This randomness of traversal also ensures that all possible combinations of equations are generated instead of a deterministic single set of all equations being included every time in the generated physics word problem. For each equation, we select one unknown variable. If we traverse from a source node to a destination node via an edge, the edge becomes the unknown variable for the source node equation. Let's say the edge $(a)$ was selected from the equation node $F = ma$, then $(a)$ becomes the unknown variable for that equation, and we put it in the set of unknown variables. Since the next equation ($S = u \cdot t + 0.5 \cdot a \cdot t^2$) node is our last node, we select a variable randomly from $(S)$, $(u)$, $(t)$ to be in our unknown set. Note, that the algorithm doesn't select variables already in the unknown set. Let's say we select $(u)$, so after the final traversal, the known set contains $\epsilon \{F, S, t, m\}$ and the unknown set contains $\epsilon \{a, u\}$, and since we have two independent and consistent equations with two unknowns, we can solve them. Finally, the \textbf{Equation Builder} module outputs a set of known and unknown variables along with their units, variable names, and values. Variables that are vectors are sometimes broken down into their components. Each topic is associated with a configuration file that includes a set of relevant equations, the names of all associated variables, their possible value ranges, units, and an indication of whether each variable is a scalar or a vector. An example of such a configuration file is present in Appendix \ref{config}. Once the Equation Builder produces the set of equations that form the basis of our physics problem, the pipeline proceeds to the Validation phase. During this stage, solvability checks are performed, such as verifying whether the equation set can be solved using Newton’s method\cite{kelley2003solving}. This validation component is optional and was not utilized in our experiments. Following validation, the workflow enters the \textbf{Topic Phrase Selection} phase, where a novel agentic Retrieval-Augmented Generation (RAG) system is used to retrieve topic-relevant concepts and phrases associated with the generated equation set. For our example, let’s say the \textbf{Topic Phrase Selector} selects a topic phrase like “car in a racetrack” and from \textbf{Equation Builder} we get something like "mass = 10 kg, initial velocity = 20 $ms^{-1}$, time = 10 sec, force = 30N, acceleration = unknown, distance = unknown". 
These structured inputs are then provided to an LLM, which converts them into a natural-language physics problem. Appendix~\ref{Ours} presents intermediate prompts generated by the \textbf{Equation Builder} and \textbf{Topic Phrase Selector} components for various topics, along with the final question produced by the LLM. An example output is: "A truck of mass 10 kg is initially moving with a velocity of 20 m/s. A constant force of 30 N is applied in the direction of motion for 10 seconds. Assuming the track is straight and friction is negligible, determine the acceleration of the truck and the distance covered in 10 seconds." A complete overview of this process is shown in Figure~\ref{fig:arch}. To increase the diversity of generated questions, an optional \textbf{Synonym Adder} can be applied before the prompt is sent to the LLM. This component replaces entities or objects in the selected phrase with semantically equivalent alternatives; for example, "car on a racetrack" may be transformed into "truck on a racetrack". Finally, the generated question can be optionally verified using an \textbf{Equation Validator} to ensure consistency between the underlying equations and the textual description. In our experiments, we utilized a trusted-editing strategy using a secondary LLM to perform this validation step. \cite{greenblatt2024aicontrolimprovingsafety}.\\
\\
Next, we briefly describe the pseudocode of our framework. Algorithm 1 is the main entry point of the \textbf{ARVRE} framework and orchestrates the overall pipeline by sequentially invoking key components: the Equation Builder, the Topic Phrase Selector, and finally the LLMs, which together generate a physics question for a given topic. Function BuildEquationSet in Algorithm 2 details the operation of the Equation Builder, which constructs known and unknown variables during inference; this component is also illustrated on the left side of Figure~\ref{fig:arch}. Additionally, the function ExtractTopicWords (Appendix~\ref{agentic_rag}) describes the Topic Phrase Selector, which leverages a novel iterative RAG-based approach to extract relevant topic words. In the following sections, we provide a detailed description of each component of our ARVRE framework.

\subsection{Equation Builder}
The \textbf{Equation Builder} component performs two key functions. First, it pre-computes an equation graph for each topic offline, before any user request is made, as described in the previous section. Second, when a user requests the generation of a physics word problem, the \textbf{Equation Builder} selects $N$ equations from the chosen topic and determines $N$ unknown variables using the graph traversal method outlined earlier. This traversal process is guided by an $\epsilon$-greedy policy over the state–action value function $Q(s,a)$, which is learned using the SARSA algorithm.

\textbf{SARSA-based Reinforcement Learning Formulation And Its Benefits:} In inference mode, our framework ARVRE selects a greedy action $argmax_a Q(s,a)$ with a probability of $1-\epsilon+\epsilon/N$. In training mode, we update state–action values using the offline SARSA update rule presented in Equation[\ref{sarsa}]. Each reinforcement learning episode is defined as the complete generation of the physics word problem from the beginning. During training the corresponding state–action pairs are updated using delayed user feedback received at the end of the episode. The user assigns a reward between 0 and 1 based on any or all metrics like solvability, novelty and problem difficulty. Instead of relying on user provided rewards, we can leverage LLMs to generate reward signals based on user-defined evaluation metrics. Since user ratings exhibit stochasticity and are not strictly consistent across interactions, we adopt SARSA \cite{rummery1994line} rather than the Q-learning update \cite{watkins1992q} (Equation [\ref{qlearning}]). SARSA is considered safer in uncertain environments because the updates are on-policy updates and are less overoptimistic(no max operator) compared to off-policy Q-learning updates. As discussed in \cite{hasselt2010double} Q-learning overestimates action value pair. 

\begin{equation}
    Q(s,a) = (1-\alpha)Q(s,a) + \alpha[r + \gamma \cdot Q(s',a')]
    \label{sarsa}
\end{equation}
\begin{equation}
    Q(s,a) = (1-\alpha)Q(s,a) + \alpha[r + max_{a'}\gamma \cdot Q(s',a')]
    \label{qlearning}
\end{equation}

In the above equations, $s$ and $s'$ represent the current and the next equation nodes, respectively, while $a$ and $a'$ represent the edges that connect equation nodes. Although the updates follow the SARSA rule in Equation~\eqref{sarsa}, the overall process is not a strict Markov Decision Process (MDP). This is because the value of the current equation node $V(s_{curr})$, depends not only on its immediate parent but also on the entire history of previously visited states. Nevertheless, we model it as a markov decision process(approximation), since it enables simple state–action value updates that work well empirically in guiding the search toward solvable equation chains. When the user provides a reward signal based on novelty and solvability, the equation set (along with the known-unknown variable) that resulted in the final generation of physics word problems gets appropriately reinforced. A bad equation set gets penalized, and vice versa. As a result, over time, our method produces word problems that are more novel and solvable. 

\subsection{Topic Phrase Selector}
We use an agentic RAG (Retrieval-Augmented Generation) framework to select relevant topic words/phrases for a given set of equations. First, we build vector embeddings of physics textbooks using a sentence-transformer model(which maps sentences into vectors inside a vector space, where semantic similarity corresponds to vector proximity), this enables efficient nearest-neighbor search for retrieval tasks \cite{reimers2019sentencebertsentenceembeddingsusing}, and we store them in a vector database (Qdrant was used because it supports both semantic and syntactic search). Then, using the reasoning capabilities of an LLM, we retrieve topic words/phrases that best match the set of equations by querying the vector database. When the LLM receives the set of equations, it is prompted to access the vector database and fetch relevant topic words/phrases. After the first iteration of topic-word retrieval, we ask the LLM to evaluate whether the topic words/phrases provide sufficient context for the given equations to build a word problem. If it deems the list inadequate, the LLM is permitted to add sentences or refine the original prompt (which already includes the set of equations) to produce a richer set of topic words/phrases. This loop of retrieval, evaluation, refinement continues until the LLM is satisfied or a specified maximum number of iterations are reached. Because of recent advances in LLM reasoning and RAG techniques, the topic phrase retrieval will get better as we switch to a modern version of the LLM or vector database. Once the topic words/phrases are finalized, they, together with the set of equations, are passed on to the next component, the Synonym Adder. In our “agentic” variant, the LLM not only consumes retrieved content but also acts on it. The LLM initiates further retrieval loops by refining the prompt or augmenting context and decides when to stop. This introduces an iterative, agent-style overlay on the traditional RAG pipeline\cite{lewis2020retrieval} \cite{gao2023retrieval}\cite{lewis2021retrievalaugmentedgenerationknowledgeintensivenlp}, which only retrieves relevant text passages once and sends them into the generative model. Appendix~\ref{FullOurs} provides an example of the topic phrases retrieved by our agentic RAG framework given the equation prompt generated by the Equation Builder component.

\subsection{Synonym Adder}
The synonym adder is an optional module designed to improve the diversity of generated physics questions by replacing certain words with their synonyms. This module utilizes WordNet’s synsets to identify synonyms, but this approach has limitations. Synsets often produce incomplete synonym lists and may replace words with their base lemmas rather than true synonyms. A more effective alternative would be to use Merriam-Webster’s thesaurus or to leverage LLMs for more contextually appropriate replacements. We experimented with all the different strategies. Using LLMs proved to be simpler and more effective. While this module is not essential to the core model, it was incorporated to produce slight variations in the generated questions.

\subsection{LLMs}
We leverage large language models (LLMs) to generate coherent sentences based on core equations and context-specific topic words/phrases produced by the previous components. To determine the most effective model for this task, we experimented with both open-source and commercial options of varying model size, including Llama-3.2-3B-Instruct, Mistral-7B-Instruct \cite{chaplot2023albert}, DeepSeek-R1, and the paid GPT-3.5-Turbo. After extensive testing, we selected GPT-3.5-Turbo as our primary LLM due to its superior performance, particularly when completing sentences for large equation sets compared to the other LLMs. We note that in future, more capable LLMs are expected to yield further improvements. In addition to using GPT-3.5-Turbo, we created our own Physics Question Dataset (\textbf{PQD}) and fine-tuned a GPT-2 model \cite{radford2019language} for question generation. This was done to assess whether a lighter, open-source model could produce sentences of comparable quality to GPT-3.5-Turbo. While fine-tuning significantly improved GPT-2’s performance, it still fell short of GPT-3.5-Turbo in generating complex and contextually accurate questions.

\subsection{Equation Validator}
We have two types of equation validators(optional). The first equation validator uses Newton’s Method to find the solution to the set of equations generated from the Equation Builder component. Let $F(x) = 0$ be a system of nonlinear equations. $F = [f_1, f_2,... f_n]^T$(vector of n equations), $x = [x_1, x_2,... x_n]^T$(vector of n unknowns), using Equation (\ref{newton_raphson}) we refine the values of the next set of unknowns.
\begin{align}
    x_{k+1} = x_{k} - J(x_k)^{-1}F(x_k)
    \label{newton_raphson}
\end{align}
Newton’s method is not perfect, it has many limitations for example, it depends on a good initial guess, evaluating the Jacobian could be time consuming and it could also be singular. We can incorporate this additional validation step after the Equation Builder module to improve solvability, where we attempt to solve the resulting nonlinear system using Newton's method \cite{kelley2003solving}. If the system is unsolvable, we prompt the Equation Builder to generate a new question. In the linear case, solvability is governed by the Rouché–Capelli theorem, which compares the rank of the coefficient matrix with that of the augmented matrix \cite{strang2006linear}. This filtering process ensures that our questions are not only novel but also solvable before they are passed to the LLM. 
We optionally introduce a second type of Validator that operates on the final response generated by the LLM. Although the generation process is guided using predefined known variables, unknown variables, and topic-specific words or phrases, the LLM is not guaranteed to strictly adhere to these constraints. To address this, we utilize an additional LLM to evaluate the generated word problem. This validator LLM is tasked with solving the generated problem using the provided set of equations. If it fails to do so, it is prompted to revise the problem so that it becomes solvable under the given equation set. Several evaluation strategies can be applied in this context, including trusted monitoring and trusted editing \cite{greenblatt2023ai}. In our work, we focus only on the trusted editing approach, where errors in the initial draft of the word problem (produced by the primary LLM) are identified and corrected by the validator LLM to ensure solvability.

\section{Properties and Implications}
The following definitions are presented to understand the proposed algorithm.

\textbf{Definition 1:} The complexity cost (CC) of a node $p_w$ for a trajectory ${\tau}$, denoted by $C_{\tau}(p_w)$, is defined as the total number of unique terms introduced along the path from the root node $r$ to the node $p_w$, where $\tau: r...,p_{w-1},p_w$ is the trajectory. The root node is a dummy node with no associated equation, which implies that $u(r)=\phi$. 
where,
$$u(k) = \text{a set of variables in equation node k}$$
   
Then,
\begin{align}
    C_{\tau}(p_w) &= u(p_w) \cup  u(p_{w-1})\cup u(p_{w-2}) ... \cup u(p_{w-k}) \cup u(r) \nonumber \\
    C_{\tau}(p_w) &= \bigcup_{i=0}^{k}u(p_{w-i})
    \label{complexity}
\end{align}

This metric serves as a proxy for the complexity of the final physics word problem. The underlying intuition is that introducing a larger number of variables generally increases the difficulty of solving the problem.

\textbf{Property 1:} The complexity cost (CC) of a node $p_w$ for a trajectory ${\tau}$, denoted by $C_{\tau}(p_w)$, always satisfies the following inequality:
\begin{equation}
C_{\tau}(p_w) \geq C_{\tau}(p_{w-1}) \geq C_{\tau}(p_{w-2}) \geq \cdots \geq C_{\tau}(p_{w-k}) \geq C_{\tau}(r)
\label{complexityCost}
\end{equation}

\textbf{Property 2:} The ARVRE algorithm is complete, i.e., if a valid equation set exists for a given topic, ARVRE will eventually discover it.

\textbf{Property 3:} Graph traversal by ARVRE while selecting equation sets is acyclic, that is, the same equation nodes are not visited twice during a single graph traversal.

\textbf{Property 4:} The method assigns a higher selection probability to the equation set, along with its known and unknown sets, if they are solvable.

\textbf{Property 5:} If our equation chain(generated during graph traversal) has \textbf{N} equations then we have \textbf{N} unknown variables.

\textbf{Property 6:} For two different runs of the ARVRE algorithm on the same topic, different equation sets can be selected.

\textbf{Property 7:} When training is enabled, the LLM progressively generates more solvable questions, given that the reward signal consistently assigns higher scores to solvable questions.

\textbf{Property 8:} Different LLMs develop distinct preferences over subsets of all equation chains.

Proofs of some of the above properties are provided in Appendix~\ref{Proof}.

\section{Results}
\begin{figure*}[t]
    \captionsetup{type=table}
    \centering
    \begin{tabular}{|c|m{7.5cm}|m{5cm}|}
        \hline
        Model Name & Word Problem & Grade \\
        \hline 
        Our Model & 
        In an aerial battle, a fighter pilot releases a bomb of 5 kg with a force of 10 N at an angle of 30° above the horizontal. The plane is accelerating at 72 m/s² at an angle of 43° with the horizontal over 778 m from ground. What are the horizontal and vertical components of the force acting on the bomb? How far will the bomb travel horizontally before hitting the ground? &
        1. Solvability: 1 (Solvable) \newline
        2. Linguistic Complexity: 2 \newline
        3. Problem Hardness: 2 \\
        \hline
        \hline 
        Llama-3.2-3B-Instruct & 
        A ball is thrown upwards from the ground with an initial velocity of 25 m/s. Assuming negligible air resistance, what is the maximum height reached after 2 seconds? &
        1. Solvability: 1 (Solvable) \newline
        2. Linguistic Complexity: 0 \newline
        3. Problem Hardness: 0 \\
        \hline
        \hline 
        DeepSeek-R1 & 
        A ball is thrown upwards with an initial velocity of 20 m/s. Find its velocity and displacement 3 seconds after launch. Assume acceleration due to gravity is 10 m/s². & 
        1. Solvability: 1 (Solvable) \newline
        2. Linguistic Complexity: 0 \newline
        3. Problem Hardness: 1 \\
        \hline 
        \hline 
        GPT-3.5-Turbo & 
        A ball is dropped from the top of a 50-meter-high building. Neglecting air resistance, assume the acceleration due to gravity is 9.8 m/s2
        (a) How long does it take for the ball to reach the ground?
        (b) What is the velocity of the ball just before it hits the ground?
        (c) If the ball rebounds with 60\% of its impact speed, how high does it rise after bouncing? & 
        1. Solvability: 1 (Solvable) \newline
        2. Linguistic Complexity: 1 \newline
        3. Problem Hardness: 1 \\
        \hline 
        \hline 
        Mistral-7B-Instruct-v0.2 & 
        A golfer swings a club, launching a ball at 30 m/s at a 45° angle. After 2 seconds, the ball bounces back up to a height of 15 meters. What is the mass of the golf ball? & 
        1. Solvability: 1 (Solvable) \newline
        2. Linguistic Complexity: 1 \newline
        3. Problem Hardness: 1 \\
        \hline
    \end{tabular}
     \caption{Shows a survey form which has one question from a particular topic (kinematics in this case) per model, along with the grade provided by an evaluator.}
     \label{fig:gform}
\end{figure*}

\begin{figure*}[t]
    \captionsetup{type=table}
    \centering
        \begin{tabular}{|c|l|c|c|c|c|}
            \hline
            \textbf{S. No.} & \textbf{Model} & \textbf{\% Solvable} & \textbf{Linguistic Complexity} & \textbf{Problem Hardness} \\
            \hline
            1 & Own Model & 91.38 & \textbf{*1.56} & \textbf{*1.75}  \\
            2 & DeepSeek-R1 & 93.79 & 0.42 & 0.55  \\
            3 & ChatGPT-3.5-Turbo & \textbf{*94.15} & 0.37 & 0.48 \\
            4 & Llama-3.2-3B-Instruct & 89.28 & 0.25 & 0.42 \\
            5 & Mistral-7B-Instruct-v0.2 & 90.14 & 0.22 & 0.24 \\
            \hline
        \end{tabular}
        \caption{Evaluation of different models based on solvability, problem hardness and linguistic complexity. The best performing model on each metric is highlighted using (*).}
        \label{tab:model_evaluation}
\end{figure*}

\begin{figure}
    \captionsetup{type=table}
    \centering
        \begin{tabular}{|l|c|c|}
            \hline
            \textbf{Model Name} & \textbf{Conformity Score} \\
            \hline
            Own Model & 0.0591 \\
            DeepSeek-R1 & 0.0761 \\
            GPT-3.5-Turbo & \textbf{*0.0289} \\
            Llama-3.2-3B-Instruct & 0.5873 \\
            Mistral-7B-Instruct-v0.2 & 0.3188 \\
            \hline
        \end{tabular}
        \caption{Conformity Scores of Various Large Language Models. A lower conformity score indicates more diverse question generation. The best model is highlighted (*).}
        \label{tab:novelty_scores}
\end{figure}

\begin{figure}
    \centering
    \includegraphics[width=1\linewidth]{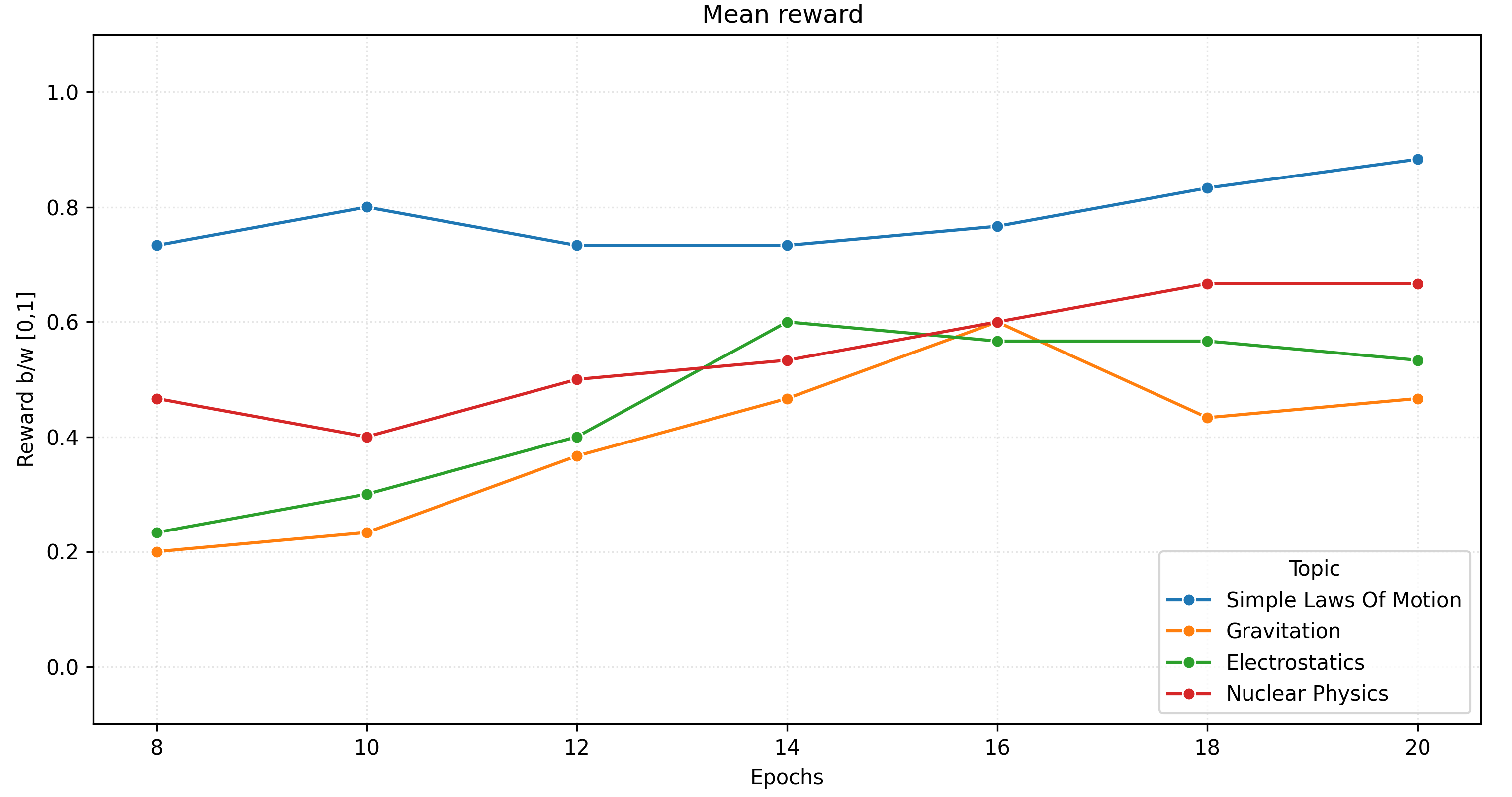}
    \caption{Mean reward per physics topic during reinforcement learning training, computed over epochs using a rolling window of size 3.}
    \label{fig:graph}
\end{figure}

To evaluate our model's performance, we compare it against several state-of-the-art language models, including Llama-3.2-3B-Instruct \cite{dubey2024llama}, Mistral-7B-Instruct \cite{chaplot2023albert}, DeepSeek-R1 \cite{guo2025deepseek}, and GPT-3.5-Turbo, all the models were asked to produce a word problem from a particular topic. Human evaluators were then used to assess the model outputs based on three key metrics: solvability, problem hardness, and linguistic complexity, as described below. Additionally, the models were also compared using the conformity score metric given by Equation \eqref{bleu_score}.

1. Solvability – Determines whether the problem is mathematically solvable. A score of 1 was assigned if the evaluator felt that the problem was solvable; otherwise, a score of 0 was assigned. If a problem was ambiguous or lacked sufficient information, it was classified as unsolvable by the evaluators.

2. Problem Hardness - Assess the conceptual and cognitive effort required to solve the problem. The scores ranged from 0 to 2, where 0 indicates a simple problem and 2 represents a highly challenging problem.

3. Linguistic complexity - Evaluates the complexity of sentence structures, vocabulary diversity, and information concealment within the problem statement. A score of \textbf{0} indicates minimal complexity, characterized by clear, straightforward wording and simple sentence structures. On the other end a score of \textbf{2} implies a high complexity, featuring sophisticated sentence constructions, abstract wording, and reliance on some implicit knowledge.

4. Conformity score - The conformity score measures how similar the current generation of Physics Word Problems (PWPs) are to previous generations. A lower score indicates greater diversity, meaning that the model is not repetitively generating similar problems. A lower conformity score is desirable because it means that the model produces novel and diverse physics problems rather than reusing existing problem structures.
{\small
\begin{align}
CS(i,j) &= \min\!\left(1, e^{1 - \frac{l_i}{l_j}}\right)
          \cdot e^{\sum_{t=1}^{N} \log p_t}, \nonumber \\
CS(j,i) &= \min\!\left(1, e^{1 - \frac{l_j}{l_i}}\right)
          \cdot e^{\sum_{t=1}^{N} \log p_t}, \nonumber \\
\text{Score} &= \frac{CS(i, j) + CS(j, i)}{2}.
\label{bleu_score}
\end{align}
}

To evaluate the effectiveness of our model, we recruited 50 undergraduate engineering students who voluntarily participated in the evaluation process. The participants remained anonymous and received no compensation. They were given clear instructions to complete the task at their own pace. All evaluators were proficient in English and physics and were familiar with the nature of the task. The participants received a Google Form that contained one question per model on the same topic. They were asked to evaluate questions based on the solvability, problem hardness, and linguistic complexity. To prevent fatigue from affecting response quality, we limited the evaluation to five questions (one from each model) without exposing which question belonged to which model. The questions in the Google Form were randomly sampled from each model's data bank, without repetition. Table [\ref{fig:gform}] presents an example of a Google Form with evaluator-assigned ratings, and Appendix \ref{Ours} presents examples of more of these generated word problems. Table [\ref{tab:model_evaluation}] summarizes model performance across various metrics. Our model excels in generating problems with high linguistic complexity and problem hardness but shows slightly lower performance in terms of solvability, despite using Equation Validators to ensure the underlying problem is valid. This discrepancy arises primarily due to two reasons. First, when the LLM is tasked with generating a word problem based on a complex equation involving many known and unknown variables, it occasionally omits some variables in the final question. We expect this to improve with advancements in LLM capabilities. Second, for more challenging problems, evaluators sometimes perceive solvable questions as unsolvable, leading to false negatives in the solvability metric.

Among the four metrics, the conformity score was calculated using Equation \eqref{bleu_score}. To compute this, we selected a pair of questions denoted by 
$i$ and $j$, from each LLM’s topic specific question bank. The BLEU score \cite{papineni2002bleu} was calculated twice—first using question $i$ as the reference and then using question $j$ as the reference. The final conformity score was the average of these two values. A high BLEU score indicates greater similarity between questions in the question bank. Equation \eqref{bleu_score} shows the BLEU score for the $i$-th and $j$-th questions, and this process is repeated for all question pairs in our dataset. The final conformity score is obtained by averaging these values. The results of this analysis are presented in Table [\ref{tab:novelty_scores}]. Our model ranks second in terms of generating diverse questions. The results suggest that topic words/phrases and the environment constraints may limit the LLM’s ability to produce questions in a broader linguistic domain. Theoretically, expanding the set of topic words/phrases and environmental variables will improve the conformity score of our approach.\\
\\
Figure~\ref{fig:graph} illustrates the learning behavior of the SARSA-based RL algorithm on equation selection across different physics topics. Each point in the figure represents the average reward computed over a sliding window of three consecutive training epochs. The average reward consistently increases as training progresses, indicating that our framework learns to generate higher-quality physics word problems over time. The x-axis represents training epochs, where each epoch consists of generating a set of equations for a selected physics topic, then constructing a corresponding physics word problem, and finally receiving a user/LLM rating between 0 and 1 based on the problem's novelty and solvability. The upward trend in average reward demonstrates SARSA's effectiveness in improving problem-generation quality through iterative feedback. From Figure~\ref{fig:graph}, we notice that topics like `Simple Laws Of Motion` start with a very high reward, indicating that the LLMs we used had an easier time generating questions on this topic.

\textbf{The Effectiveness of Our Method:} Simply prompting the LLM to generate physics problems tends to produce word problems with standard levels of \textbf{conformity}, \textbf{problem hardness}, and \textbf{linguistic complexity}—reflecting the patterns on which the model was trained. A basic prompt leads the model to generate the most likely tokens(depending on sampling parameters in the final layer of the LLM), resulting in typical problem types. However, by designing an underlying prompt structure with a valid set of known and unknown variables, we actively guide the LLM to generate more novel and solvable physics word problems.

\section{Conclusion}
Our proposed algorithm successfully addresses the challenge of generating novel, complex, and solvable Physics Word Problems(PWPs) that cannot be produced by standard prompts given to off-the-shelf large language models(LLMs). By starting with a valid set of equations and chaining them into large, intricate systems, our method surpasses the capabilities of existing LLMs in producing sophisticated and diverse physics questions. Our method also introduces a tunable threshold parameter $(TH)$, which can be adjusted to change the difficulty of the generated physics questions, our method also uses reinforcement learning to select equation sets. If a given LLM fails to generate a valid physics problem from a particular set of equations, the RL mechanism deprioritizes that set, reducing its likelihood of being selected again. In future work, we aim to expand our generator to cover a broader range of Physics domains, such as thermodynamics, electromagnetism, and quantum mechanics, with more complicated equation systems and constraints. To improve the validity of the generated questions due to occasional errors made by the LLM, we are planning to fine-tune the LLMs using low-rank adapters for each physics topic.

\appendices
\onecolumn

\section{Configuration file example for a topic.}\label{config}
Below is the configuration file for "Simple Laws of Motion", as an example. I will explain some of the key elements in the file. For instance, \textbf{\{"topic": "simple laws of motion"\}} specifies the name of the topic, \textbf{\{"equations": [...]\}} contains the list of equations associated with the topic, and\textbf{\{"variable\_names": \{...\}\}} provides detailed information about each variable. This includes the variable's name, its range of possible values, the unit of measurement, whether it is a scalar or a vector, and whether it accepts real or integer values.

\begin{multicols}{2}
\begin{verbatim}
{
    "topic": "simple laws of motion",
    "equations": [
        "v = u + a * t",
        "S = u * t + 0.5 * a * t ^ 2",
        "F = m * a"
    ],
    "variable_names": {
        "v": [
            "final velocity",
            [0, 100],
            "m/s",
            "V",
            "R"
        ],
        "u": [
            "initial velocity",
            [0, 100],
            "m/s",
            "V",
            "R"
        ],
        "a": [
            "acceleration",
            [0, 100],
            "m/s²",
            "V",
            "R"
        ],
        "t": [
            "time",
            [1, 100],
            "s",
            "S",
            "R"
        ],
        "S": [
            "displacement",
            [0, 100],
            "m",
            "V",
            "R"
        ],
        "g": [
            "acceleration due to 
            gravity",
            10,
            "m/s²",
            "V",
            "R"
        ],
        "F": [
            "force",
            [1, 100],
            "N",
            "V",
            "R"
        ],
        "m": [
            "mass",
            [1, 100],
            "kg",
            "S",
            "R"
        ]
    }
}
\end{verbatim}
\end{multicols}

For the topic "Simple Laws of Motion," the corresponding environment configuration file (shown below) is presented. Each configuration file contains multiple environment elements. In this example, three environments: \textsl{train, cave,} and \textsl{ground} are included.
Each environment element begins with the name of the environment and may also include additional properties. For example, the \textsl{train} environment includes dimensional properties, such as length, and can also have associated topic words. When a topic is selected by the topic selector, one environment is chosen along with one associated topic word, both at random. The structure of any environment properties (such as a train's length), if present, follows the same pattern as the variables of equations defined in the topic configuration file(above).
\clearpage

\section{Topic Phrase Selector}\label{agentic_rag}
The algorithm below describes how, given an equation prompt $P_{eqn}$, we use the reasoning capabilities of LLMs to extract relevant topic words/phrases from a document repository using semantic retrieval (RAG). The LLM iteratively controls the prompt $P$ to retrieve additional information when needed, and decides when sufficient context has been gathered for question generation. This iterative retrieval and stopping mechanism is captured in Lines 7–8 of the algorithm.
\begin{minipage}{1.0\textwidth}
\begin{algorithm}[H]
\caption{Topic Word Extraction from Equation Prompts}
\begin{algorithmic}[1]
\Function{ExtractTopicWords}{$P_{eqn}$}
\State Set max iterations $M \gets 5$, $t \gets 0$
\State Initialize topic words $T \gets \emptyset$, prompt $P \gets P_{eqn}$
\While{$t < M$}
    \State Retrieve related words $W \gets \text{VectorDB}(P)$
    \State Update topic words: $T \gets T \cup W$
    \State Query LLM with $(T, P_{eqn}, P)$ to obtain halt signal $H$
    \State Query LLM with $(T, P_{eqn}, P)$ to get new control prompt $P'$
    \If{$H$ is true}
        \State \textbf{break}
    \EndIf
    \State Update prompt: $P \gets P'$
    \State Increment iteration: $t \gets t + 1$
\EndWhile
\State \Return $T$
\EndFunction
\end{algorithmic}
\end{algorithm}
\end{minipage}

\section{Some questions generated by our method.}\label{Ours} 
\begin{figure}[h]
\centering
\includegraphics[width=1.05\textwidth]{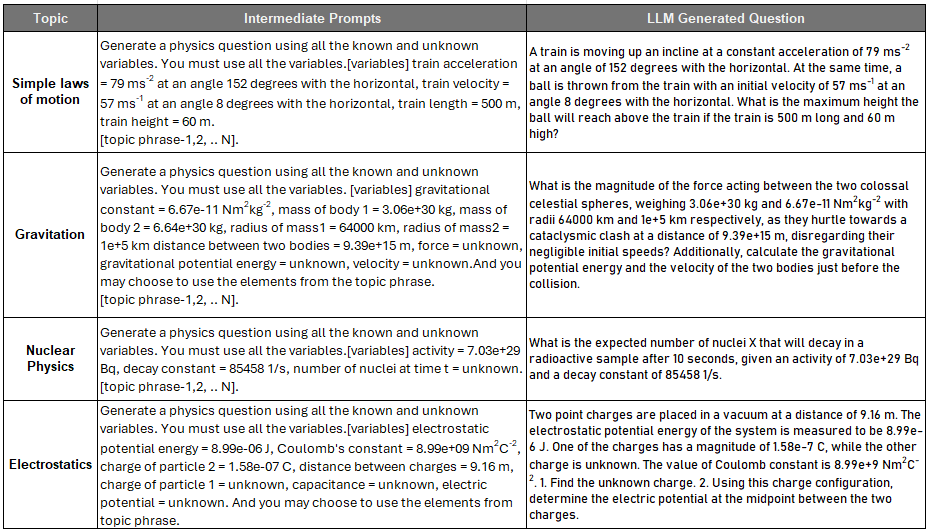}
\caption{Examples of topics, intermediate prompts, and LLM-generated final response from left to right column, using our approach. Topic phrases in intermediate prompts are redacted.}
\end{figure}
\clearpage

\section{Topic phrase generated by agentic RAG}\label{FullOurs}
\begin{figure}[h]
\centering
\includegraphics[width=1.05\textwidth]{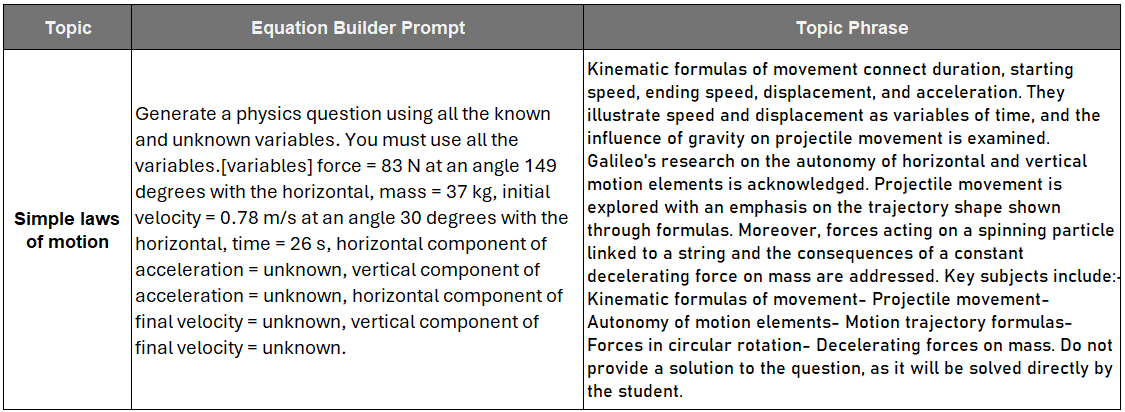}
\caption{Examples of topics, intermediate prompts, and LLM-generated final response from left to right column, using our approach. Topic phrases in intermediate prompts are redacted.}
\end{figure}

\textbf{Final generated physics problem:} A mass of 37 kg is subjected to a force of 83 N acting at an angle of 149 degrees with the horizontal for a duration of 26 seconds. Initially, the mass is moving at a velocity of 0.78 m/s at an angle of 30 degrees with the horizontal. 1. Decompose the initial velocity into its horizontal (\(v_{0x}\)) and vertical (\(v_{0y}\)) components.
2. Determine the horizontal and vertical components of the applied force.
3. Using the applied force, calculate the horizontal and vertical components of acceleration (\(a_x\) and \(a_y\)) acting on the mass. 4. Using kinematic equations, derive the horizontal component of final velocity (\(v_{fx}\)) and vertical component of final velocity (\(v_{fy}\)) after 26 seconds, including the influence of gravity on vertical motion. Your task is to find:\\
a. The horizontal component of acceleration (\(a_x\))\\
b. The vertical component of acceleration (\(a_y\))\\
c. The horizontal component of final velocity (\(v_{fx}\))\\
d. The vertical component of final velocity (\(v_{fy}\))\\
Make sure to represent all quantities using the appropriate unit vectors where applicable, and illustrate how time influences the motion via the kinematic formulas.
\clearpage

\twocolumn
\section{Proof of the common properties of the ARVRE algorithm}\label{Proof}

\textbf{Property 1:} The complexity cost (CC) of a node $p_w$ for a trajectory ${\tau}$, denoted by $C_{\tau}(p_w)$, always satisfies the following inequality:
\begin{equation}
C_{\tau}(p_w) \geq C_{\tau}(p_{w-1}) \geq C_{\tau}(p_{w-2}) \geq \cdots \geq C_{\tau}(p_{w-k}) \geq C_{\tau}(r)
\label{complexityCost}
\end{equation}

We can write the complexity cost of a node $q_w$ using the equation [\ref{complexityX}] below.

\begin{align}
    C_{\tau}(p_w) &= u(p_w) \cup  u(p_{w-1})\cup u(p_{w-2}) ... \cup u(p_{w-k}) \cup u(r) \label{complexityX} \\
    C_{\tau}(p_w) &= u(p_w) \cup \bigcup_{i=1}^{k}u(p_{w-i}) \nonumber \\
    C_{\tau}(p_w) &= u(p_w) \cup C_{\tau}(p_{w-1}) \nonumber \\
    |C_{\tau}(p_w)| &= |u(p_w)| + |C_{\tau}(p_{w-1})| - |u(p_w) \cap C_{\tau}(p_{w-1})| \nonumber \\
    |C_{\tau}(p_w)| &\geq |C_{\tau}(p_{w-1})| \nonumber
\end{align}

From induction we can derive the inequality in equation [\ref{complexityCost}], therefore as we add more equation nodes from the root the complexity cost of the trajectory/problem increases.\\

\textbf{Property 2:} The ARVRE algorithm is complete, i.e., if a valid equation set exists for a given topic, ARVRE will eventually discover it. \\
\emph{Proof:} Let $\tau: r...,p_{w-1},p_w$ be a valid equation chain. If we start from root node $r$ we can always trace this path because we are using $\epsilon$-greedy algorithm, every feasible action (i.e. choice of the next equation node) has a non-zero probability of being selected at each step, therefore the probability of $\tau$ is strictly positive and over sufficiently many episodes, ARVRE will eventually sample this path. Once $\tau$ is discovered, its validity yields a higher reward signal compared to invalid chains. The reinforcement learning updates therefore increase the preference for the state–action pairs along $\tau$, making the algorithm increasingly likely to reselect this path in future episodes. Hence, any valid equation chain that exists in the search space will eventually be found, which establishes the completeness of the ARVRE algorithm. \\

\textbf{Property 3:} Graph traversal by ARVRE while selecting equation sets is acyclic, that is, the same equation nodes are not visited twice during a single graph traversal. \\
\emph{Proof:} Let $S_v$ denote the set of equation nodes visited so far. According to the ARVRE algorithm, any node that already belongs to $S_v$ is not expanded further. This constraint prevents revisiting the same equation node multiple times during traversal, thereby ensuring that each node is expanded at most once and eliminating the possibility of redundant expansions or cycles. \\ 

\textbf{Property 4:} The method assigns a higher selection probability to the equation set, along with its known and unknown sets, if they are solvable. \\
\emph{Proof:} The value of each transition between equation nodes is updated using the SARSA update rule in Equation~\eqref{sarsa}. From a current equation node $s$, an action $a$ selects the next node $s'$. If $s'$ corresponds to an unsolvable or low-quality state, the resulting reward $r$ (and subsequent return) will be low or negative, which decreases the corresponding action value $Q(s,a)$. As a result, during inference, the transition (edge) associated with action $a$ becomes less likely to be selected. Conversely, transitions that lead to solvable equation chains receive consistently higher rewards, causing the associated $Q$-values to increase over training. Under the $\epsilon$-greedy policy, actions with higher $Q$-values are selected with higher probability during exploitation. Therefore, the method progressively biases the policy toward equation chains that are ultimately solvable, assigning them higher selection probabilities over time. \\

\textbf{Property 5:} If our equation chain(generated during graph traversal) has \textbf{N} equations then we have \textbf{N} unknown variables. \\
\emph{Proof:} As per the ARVRE algorithm the equation nodes are connected by edges, which represent a common variable between the nodes in the graph, this edge becomes an unknown variable. So if there are $N$ connected nodes we will have $N-1$ edges that from a chain. For the terminal equation node, an additional variable is selected from the equation that has not appeared earlier in the unknown-variable set. Therefore, the total number of selected unknown variables becomes \textbf{N}. Hence, for a chain consisting of \textbf{N} equation nodes, the construction yields exactly \textbf{N} unknown variables. \\

\textbf{Property 6:} For two different runs of the ARVRE algorithm on the same topic, different equation sets can be selected. \\
\emph{Proof:} ARVRE uses $\epsilon$-greedy exploration strategy when selecting transitions between equation nodes. Even after multiple episodes of training, the exploration rate $\epsilon$ doesn’t become zero but is bounded by a lower threshold $\epsilon_{\min}$. As a result, there remains a non-zero probability of selecting suboptimal or alternative actions at each step. Hence, the algorithm does not collapse to a single deterministic chain, and different runs of ARVRE can yield different equation chains.\\

\textbf{Property 7:} When training is enabled, the LLM progressively generates more solvable questions, given that the reward signal consistently assigns higher scores to solvable questions. \\
\emph{Proof:} Same as \textbf{Property 4:} \\

\textbf{Property 8:} Different LLMs develop distinct preferences over subsets of equation chains. \\ 
\emph{Proof:} Let $LLM_1(\alpha)$ \& $LLM_2(\beta)$ be two large language models, parameterized by model weights $\alpha$ \& $\beta$. Due to differences in pretraining data, optimization trajectories, etc, the models assign different likelihoods to questions derived from the same underlying equation chains. Reinforcement learning then increases the likelihood of selecting equation sets that lead to better questions being generated for different LLMs.

\clearpage

\ifCLASSOPTIONcaptionsoff
  \newpage
\fi

\bibliographystyle{IEEEtran}
\bibliography{ref}

\end{document}